\documentclass[10pt,twocolumn,letterpaper]{article}

\usepackage{cvpr}              
\usepackage{times}  
\usepackage{helvet}  
\usepackage{courier}  
\usepackage[hyphens]{url}  
\usepackage{graphicx} 
\usepackage{natbib}  
\usepackage{caption} 
\usepackage{multirow}
\usepackage{makecell}
\usepackage{booktabs}
\usepackage{amsmath}
\usepackage{pifont}
\usepackage{threeparttable}
\usepackage{algorithm}
\usepackage{algorithmic}

\usepackage{newfloat}
\usepackage{listings}
\usepackage[dvipsnames]{xcolor}

\definecolor{cvprblue}{rgb}{0.21,0.49,0.74}
\usepackage[pagebackref,breaklinks,colorlinks,citecolor=cvprblue]{hyperref}

\def\paperID{13530} 
\def\confName{CVPR}
\def\confYear{2024}

\title{Beyond Visual Cues: Synchronously Exploring Target-Centric Semantics for Vision-Language Tracking}

\author{Jiawei Ge, Xiangmei Chen, Jiuxin Cao\thanks{Corresponding author}, Xuelin Zhu, Bo Liu\\
Southeast University\\
{\tt\small \{jiawei\_ge, cxm\_irene, jx.cao, zhuxuelin, bliu\}@seu.edu.cn}
}

\begin{document}
\maketitle

\begin{abstract}
Single object tracking aims to locate one specific target in video sequences, given its initial state. Classical trackers rely solely on visual cues, restricting their ability to handle challenges such as appearance variations, ambiguity, and distractions. Hence, Vision-Language (VL) tracking has emerged as a promising approach, incorporating language descriptions to directly provide high-level semantics and enhance tracking performance. However, current VL trackers have not fully exploited the power of VL learning, as they suffer from limitations such as heavily relying on off-the-shelf backbones for feature extraction, ineffective VL fusion designs, and the absence of VL-related loss functions. Consequently, we present a novel tracker that progressively explores target-centric semantics for VL tracking. Specifically, we propose the first Synchronous Learning Backbone (SLB) for VL tracking, which consists of two novel modules: the Target Enhance Module (TEM) and the Semantic Aware Module (SAM). These modules enable the tracker to perceive target-related semantics and comprehend the context of both visual and textual modalities at the same pace, facilitating VL feature extraction and fusion at different semantic levels. Moreover, we devise the dense matching loss to further strengthen multi-modal representation learning. Extensive experiments on VL tracking datasets demonstrate the superiority and effectiveness of our methods.
\end{abstract}

\section{Introduction}

Over the past decades, single object tracking has been one of the most fundamental and challenging tasks in the field of computer vision. The primary objective of this task is to precisely locate a specified object (i.e., the target) within video frames, dubbed \textit{search images}. The localization process relies on the initial state of the target, which is cropped from the first frame and referred to as the \textit{template image}. Despite remarkable progress \cite{yan2021learning,chen2021transformer,li2022tracking,yan2022towards,yan2023universal,zhu2023cross,ma2022unified} has been made, these methods only leverage visual cues for tracking, neglecting the direct incorporation of high-level semantics such as \textit{categories}, \textit{attributes}, and \textit{interactions in the scene} to reveal the underlying essence of the target. Therefore, inherent challenges \cite{javed2022visual,marvasti2021deep,fiaz2019handcrafted} such as appearance variations, ambiguity, and distractions from similar objects, still persist. 

To overcome these obstacles via the power of multi-modal learning\cite{zhang2023efficient,kang2022robust,xu2023multimodal,khattak2023maple,radford2021learning,li2022blip}, the task of \textbf{Vision-Language (VL) Tracking} (or Tracking with Natural Language) \cite{li2017tracking,feng2021siamese,wang2021towards} was proposed. Unlike classical tracking methods, this task incorporates natural language descriptions as auxiliary information to specify the target. By combining the bounding box and natural language description, this approach provides explicit semantics of the target, enabling the tracker to exploit corresponding relations between visual and textual modalities. Thus, it mitigates aforementioned challenges and significantly enhances robustness and accuracy of the tracking model \cite{guodivert}.

\begin{figure}
\centering 
\includegraphics[height = 6cm]{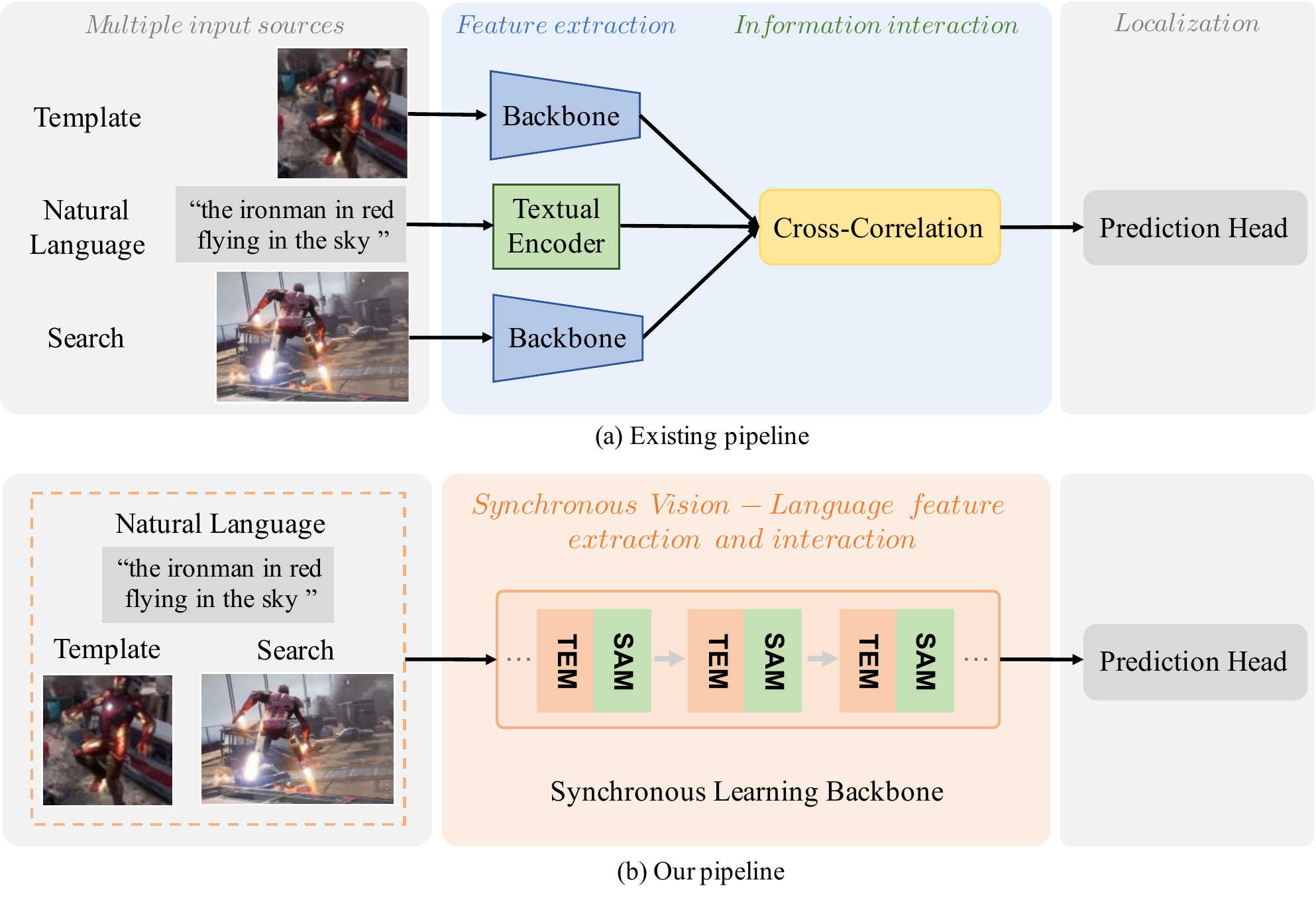}
\caption{The pipeline of existing VL trackers (a) and ours (b). We boost the performance and improve the tracking pipeline via synchronous Vision-Language feature extraction and interaction.}
\label{f1}
\end{figure}

Existing VL trackers \cite{zhou2023joint,feng2021siamese} generally comprise three components in turn (shown in Fig. \ref{f1} (a)): the backbone for feature extraction, the cross-correlation module for information interaction and the prediction head for localization. However, upon a thorough investigation of existing methods, several common drawbacks have come to light: (1) Since their backbones are originally designed for images classification, the template and search branches are unable to directly perceive each other. This absence of straight cross-correlation leads to the accumulation of target-irrelevant information, hindering the tracker from learning discriminative features suitable for the tracking task. (2) Naive cross-modal fusion designs may not effectively capture the crucial interactions among different sources (i.e. template images, search images, and natural language descriptions) at different semantic levels, which significantly compromises the benefits of multi-modal learning; (3) None of them design a proper loss function to directly optimize the multi-modal feature learning, hence limiting the model's capacity to comprehend the semantics of the target.

To overcome aforementioned issues, we improve the existing pipeline by synchronizing Vision-Language learning and information interaction from the beginning, with one single backbone network (shown in Fig. \ref{f1} (b)). This synchronous learning framework offers several advantages for VL tracking: Firstly, it effectively alleviates the accumulation of target-irrelevant noise during feature extraction by utilizing both visual and textual cues to guide representation learning right from the very beginning. This approach is inspired by the observation that \textbf{children develop recognition skills most effectively when learning visual and textual information at a similar pace \cite{smith2003learning}.} Secondly, it facilitates the capture of relationships between multiple sources at different semantic levels. Additionally, with the incorporation of an appropriate loss function, we can better unleash the potential of VL learning in the realm of tracking.

To this end, we introduce SATracker (Semantic-Aware Tracker), a VL Tracker that progressively enhances target-related features with the guidance of natural language. Specifically, we introduce two effective modules inside the Synchronous Learning Backbone (SLB) to inject target-related prior information during feature extraction: the Target Enhance Module (TEM) and Semantic Aware Module (SAM). The TEM significantly strengthens the tracker's ability to distinguish the target from the background by modifying the Multi-Head Attention \cite{vaswani2017attention} in an asymmetrical way. Through this adjustment, SATracker reinforces the representation of target-related features while effectively suppressing irrelevant background information; In SAM, we utilize the natural language description to encourage the tracker to focus on target's semantics and comprehend the context of both visual and textual modalities, enhancing the semantic-guided features. Furthermore, inspired by the concept of vision-language contrastive learning \cite{radford2021learning}, we propose the dense matching (DM) loss, which compels the tracker to exploit the target's semantics to improve performance of the VL tracking. 

In general, our contributions are four-fold:
\begin{itemize}
    \item To the best of our knowledge, we present the first Synchronous Learning Backbone (SLB) for VL tracking, which allows comprehensive information fusion across multiple sources at different semantic levels.
    
    \item The proposed Target Enhance Module (TEM) and Semantic Aware Module (SAM) are progressively integrated within the backbone, enhancing target-centric representations and perceiving visual and textual context.

    \item To fully harness the capabilities of VL representation learning, we employ the dense matching loss, which is the first VL-related loss designed for directly optimizing multi-modal learning in VL Tracking.

    \item Our approach achieves superior performance compared to state-of-the-art trackers on mainstream VL tracking datasets, demonstrating the effectiveness of our proposed method.
\end{itemize}

\section{Related Work}
\subsection{Visual Tracking} 

Siamese-based visual trackers \cite{guolearning,zhang2021learn,mayer2022transforming,zhao2022tracking,gao2022aiatrack,lin2022swintrack} have gained significant attention due to their remarkable performance on benchmark datasets. They typically employ a shared-weight backbone to extract features from both template and search regions and then locate the target by measuring the similarity between them. Recently, TransT \cite{chen2021transformer} proposed a transformer-based feature fusion network that incorporates self-attention and cross-attention mechanisms, facilitating effective feature interactions. TrDiMP \cite{wang2021transformer} further leveraged the power of transformers by utilizing the transformer encoder for feature enhancement and the transformer decoder to propagate tracking cues. More recently, MixFormer \cite{cui2022mixformer} implemented a one-stream tracking pipeline based on the CvT \cite{wu2021cvt}, achieving outstanding tracking performance. ARTrack \cite{wei2023autoregressive} transforms the tracking task into a sequence interpretation challenge, where it progressively predicts object trajectories. Despite improvements, they rely solely on visual cues, vulnerable to challenges such as ambiguity and distractions from similar objects. Therefore, there is a growing need to directly incorporate high-level semantics, such as object categories and attributes, through natural language prompt. 
\subsection{Vision-Language Tracking} 

The integration of vision and language presents promising opportunities for enhancing the robustness and addressing inherent limitations in vision tasks \cite{radford2021learning}, including tracking \cite{yang2020grounding,feng2020real}. Li et al. \cite{li2017tracking} defined the task of VL tracking, laying the foundation for subsequent research. Ma et al. \cite{ma2021capsule} introduced the CapsuleTNL, which utilizes the Capsule Network and two interaction routing modules to effectively integrate image and text features. Wang et al. \cite{wang2021towards} introduced TNL2K, a new large-scale benchmark for VL tracking. SNLT \cite{feng2021siamese} improved tracking performance with their Dynamic Aggregation Module to combine both modalities. JointNLT \cite{zhou2023joint} proposed a novel framework that reformulates grounding and tracking as a unified task. Nonetheless, existing works rely on separate visual and textual encoders to extract features, without direct interactions between modalities during representation learning. DecoupleTNL \cite{ma2023tracking} delved into two jointly optimized tasks to address the inconsistency between visual and linguistic representations. MMTrack \cite{zheng2023towards} was developed based on the ARTrack \cite{wei2023autoregressive} via incorporating language representation. In contrast, Guo et al. \cite{guodivert} introduced ModaMixer, a novel module that aims to learn a unified visual-language representation for tracking. However, it only regards language features as a selector to re-weight extracted vision features, overlooking direct interactions between the template and search features at different semantic levels, and neglecting the synchronicity of learning both modalities from the outset. All-in-One \cite{zhang2023all} maintained the ModaMixer design but replaces the backbone of VLTTT \cite{guodivert} with transformer layers. Motivated by these observations, we propose a novel framework where VL feature learning and interaction are synchronized into one single backbone. Moreover, our dense matching loss is the first VL-related loss designed for directly optimizing multi-modal representation learning in VL tracking.

\begin{figure*}
\centering 
\includegraphics[width = 17.4cm]{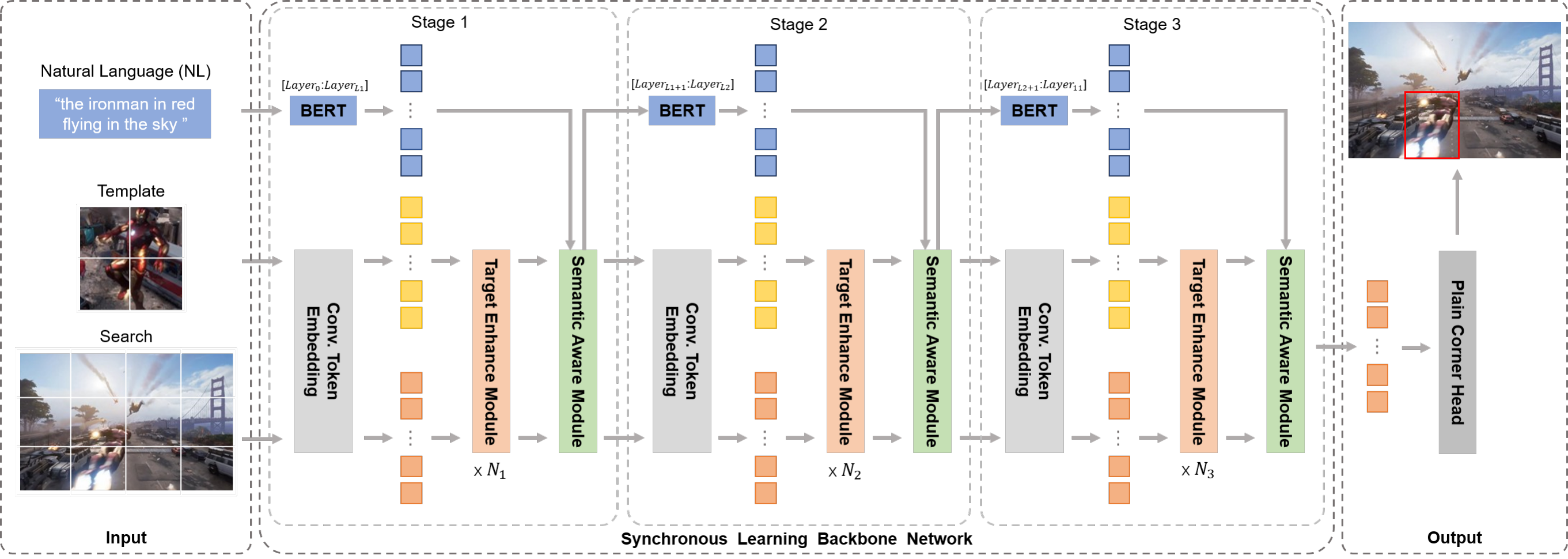}
\caption{Architecture of the proposed tracking framework. Both the template and search regions, along with the language description, are tokenized into sequences, which are subsequently sent into the synchronous learning backbone. Via the incorporation of TEM and SAM, the backbone network progressively facilitates synchronous feature extraction and interaction between the visual and textual modalities. Finally, the semantic-guided search feature is utilized for target localization through a plain corner prediction head.}
\label{f2}
\end{figure*}

\section{Method}
In this section, we present the SATracker (Semantic-Aware Tracker), a simple yet effective framework that strengthens semantics of the target from both modalities (shown in Fig. \ref{f2}). Firstly, we introduce a novel Synchronous Learning Backbone (SLB) for VL tracking, which iteratively incorporates two essential modules: the Target Enhance Module (TEM) and Semantic Aware Module (SAM). The enhanced search region features are then fed into a plain corner head to accurately predict the target location. To further promote VL representation learning, we utilise the dense matching loss, encouraging the tracker to perceive semantics of the target.

\subsection{The Synchronous Learning Backbone Network}
In our SATracker framework, the Synchronous Learning Backbone (SLB) network is crucial in modeling complex relations among various sources of visual and textual modalities. By stacking layers of Convolutional Token Embedding (CTE), Target Enhance Modules (TEM), and Semantic Aware Modules (SAM), the progressively downsampling architecture captures inductive bias similar to CNNs while forming a simple backbone that facilitates the fusion of visual and textual information to capture target-centric semantics.

Formally, the $i$-th stage ($i \in \{1,2,3\}$) takes the natural language description $T^i \in {R}^{L^i \times D^i}$ and a pair of image features (the template $X_t^i \in {R}^{H_t^i \times W_t^i \times C^i}$ and search $X_s^i \in {R}^{H_s^i \times W_s^i \times C^i}$) as inputs. Here $C^i$ denotes the channel number of the visual feature in the $i$-th stage, $H_t^i$ and $W_t^i$ represent the height and width of the image feature accordingly, $L^i$ and $D^i$ are for the length and dimension of the textual description. We first embed them using BERT and CTE. Then we feed them into the TEM and SEM to enhance target-related features and interact with each other under the guidance of high-level semantics. Finally, we utilize the output search feature of the last stage for target localization. The details of these modules are as follows. 

\subsubsection{Convolutional Token Embedding (CTE)}
This module employs a 2D convolution operation to transform the input features into tokens, providing flexibility in adjusting the dimension and reducing the number of tokens at each stage. Such downsampling architecture introduces locality and captures the inductive bias, similar to CNNs.

In stage $i$, given the image ($H^{0} \times W^{0} \times 3$) or the feature from the previous stage ($i-1$), denoted as $X^{i-1}$ with size of $H^{i-1} \times W^{i-1} \times C^{i-1}$ , we map it into $X^{i}$ via convolution and flatten them into size of $H^i W^i \times C^i$. For more details please refer to the content of CvT \cite{wu2021cvt}.

\subsubsection{Target Enhance Module (TEM)}
\begin{figure}
\centering 
\includegraphics[width = 8cm]{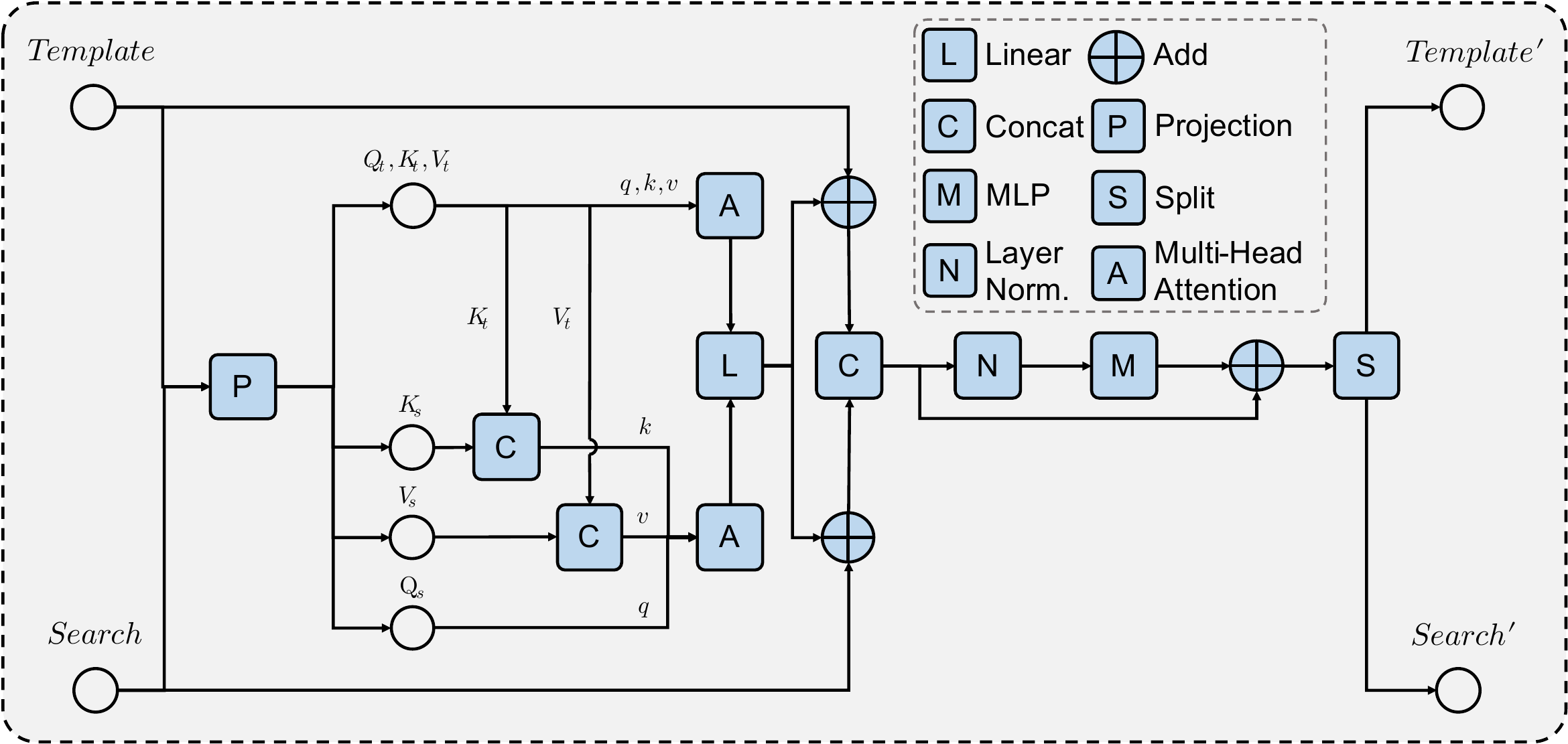}
\caption{An elaborate illustration of the Target Enhance Module, which efficiently performs self-attention and asymmetrical cross-attention to enhance target-relevant features.}
\label{f3}
\end{figure}
The Target Enhance Module (TEM) aims to extract visual features from the template and search regions while facilitating their interaction. Furthermore, to mitigate negative influence from noise in the search region on the template, we modify the conventional Multi-Head Attention to work in an asymmetrical manner. 

The detailed computation procedure is presented in Fig. \ref{f3}. We begin by reshaping the flattened features into 2D feature maps. Following CvT, we further enhance the modeling capability by applying a separable depth-wise convolutional projection on each feature map. This approach captures local spatial information and improves computational efficiency by enabling down-sampling in the key and value matrices. Subsequently, we flatten each feature map of the target template and search regions and process them through a linear projection, where \textit{queries, keys, and values} are generated and concatenated for the attention operation. We denote the template with $Q_t,K_t,V_t$ and search with $Q_s,K_s,V_s$. The concatenated \textit{key and value} are formed as $K_c= Concat(K_t, K_s), V_c= Concat(V_t, V_s)$.

Afterwards, we simultaneously perform the self-attention and asymmetrical cross-attention in an efficient manner. Here $C$ denotes the dimension of the $key$, and $Atten_t^{TEM}$ and $Atten_s^{TEM}$ represent the attention maps corresponding to the target and search regions in the TEM, respectively:

\begin{equation}
    \begin{aligned}
        Atten_t^{TEM}&={Softmax}\left(\frac{Q_t K_t^T}{\sqrt{C}}\right) V_t \\
        &= SA_t^{TEM}V_t, \\
        Atten_s^{TEM}&={Softmax}\left(\frac{Q_s K_c^T}{\sqrt{C}}\right) V_c \\
        &=Concat(SA_s^{TEM}, CA_s^{TEM})V_c,
    \end{aligned}
\end{equation}
where
\begin{equation}
    \begin{aligned}
        SA_s^{TEM} & = {Softmax}\left(\frac{Q_s K_s^T}{\sqrt{C}}\right), \\ 
        CA_s^{TEM} & = {Softmax}\left(\frac{Q_s K_t^T}{\sqrt{C}}\right).
    \end{aligned}
\end{equation}

The attention maps are then passed through a linear layer and added to their respective original tokens using a residual connection. Subsequently, the concatenated tokens are processed through a Layer Normalization and a Multilayer Perceptron (MLP), followed by another residual connection. Finally, the tokens are split for vision-language fusion within the Semantic Aware Module (SAM).

The integration of self-attention and cross-attention mechanisms within TEM empowers the module to extract comprehensive visual features and facilitates direct perception of the target, thereby enhancing the representation of target-related information.

\subsubsection{Semantic Aware Module (SAM)}
Inspired by the design of the channel-wise attention \cite{hu2018squeeze} commonly used in vision tasks, we have devised the Semantic Aware Module (SAM) to capture complex relations between visual and textual modalities. SAM aims to incorporate textual guidance into visual features and fuse visual prompt into textual features, resulting in semantic-enriched and context-aware representation.

\begin{figure}
\centering 
\includegraphics[width = 8cm]{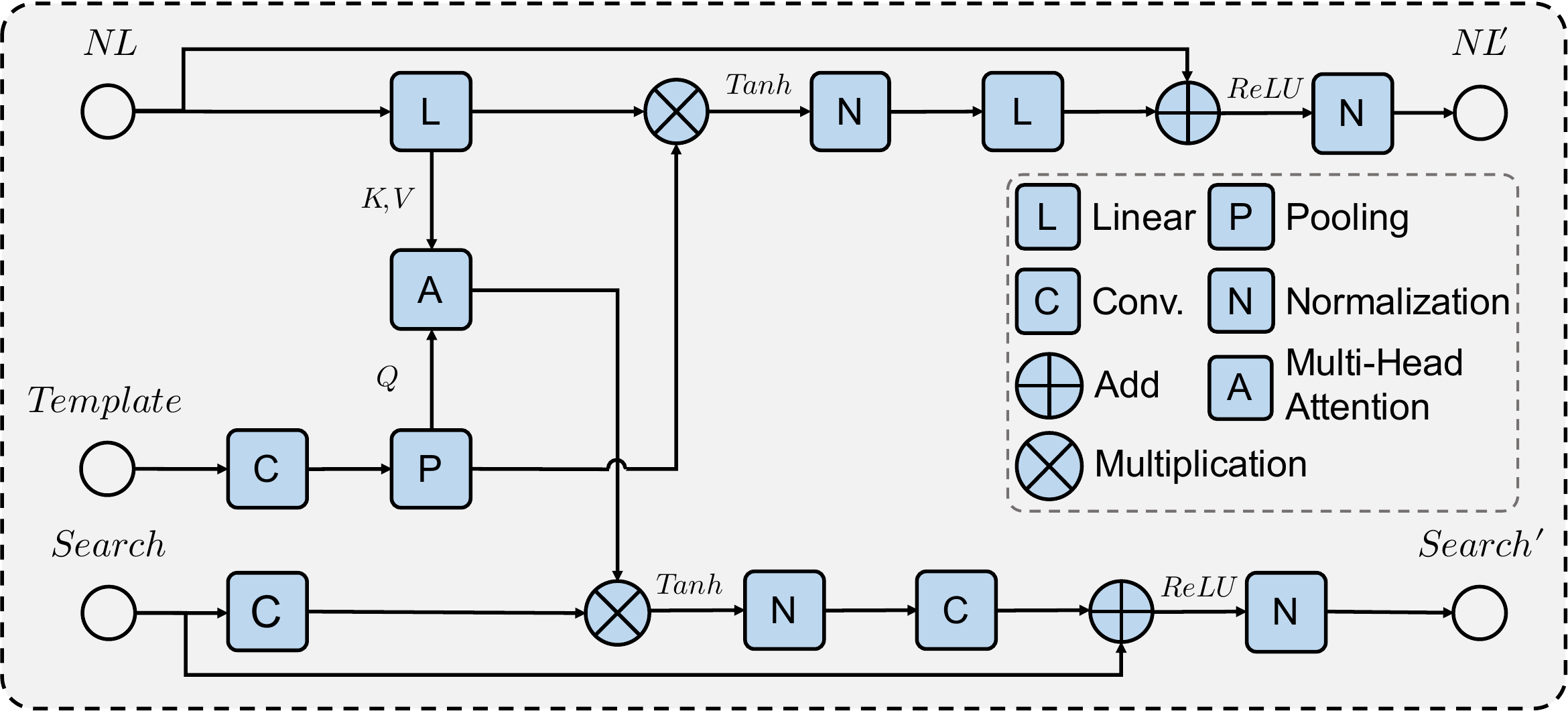}
\caption{The detailed computation process of the Semantic Aware Module (SAM), comprising the visual stream (lower side) and textual stream (upper side).}
\label{f4}
\end{figure}

In the visual stream of SAM shown in Fig. \ref{f4}, we leverage the textual feature as a weighted kernel to engage in channel-wise interaction with the visual feature. This interaction highlights visual feature maps that contain relevant context, encouraging the refined visual feature to attend to semantics of the target. Initially, the visual and textual features are mapped into a common embedding space using Convolution and Linear functions, respectively. Next, Max Pooling is applied to the template, extracting \textit{query, key, and value} for the attention operation: $Q_t$ represents the template, and $K_{NL}$ and $V_{NL}$ correspond to the Natural Language. The attention map of the Natural Language in SAM, denoted as $Atten_{NL}^{SAM}$, is generated through Multi-Head Attention:
\begin{equation}
    \begin{aligned}
        Atten_{NL}^{SAM}&={Softmax}\left(\frac{Q_t K_{NL}^T}{\sqrt{C}}\right) V_{NL}.
    \end{aligned}
\end{equation}
Here $C$ denotes the dimension of the $key$. Subsequently, we utilize $Atten_{NL}^{SAM}$ as a soft channel selector for the search region, promoting its awareness of target semantics. After Batch Normalization, we employ another CNN to ensure compatibility for the subsequent residual connection.

In the textual stream, the template directly serves as a prompt to enrich the language understanding. Concretely, the encoded natural language description is first projected using a Linear operation. Then, the pooled template feature is directly fused with the textual feature, enhancing its representation and enabling a more robust understanding of the visual context. Subsequently, the fused feature undergoes another projection through a Linear layer, followed by the residual connection to update the textual feature. This mechanism ensures that the textual feature becomes more aligned with the visual cues, enabling a deeper integration of visual and textual modalities. 

\subsection{Dense Matching Loss}
Motivated by the contrastive learning framework in CLIP, we introduce an effective vision-language loss to optimize representation learning in VL tracking. Our approach converts the image-text matching task into a dense matching problem. We believe that the region of the target should be strongly correlated with the textual description. Thus, we compute the matching score using the semantic-enhanced search feature $X_s^3$ and the target-aware text feature $T^3$ from SAM in the final stage. We also employ upsampling of the matching region to generate more matching samples, equivalent to the concept of adopting large batch size in contrastive learning:
\begin{equation}
    score= upsampling({X_s^3}^{'} {{T^3}^{'}}^{\top}).
\end{equation}
Here ${X_s^3}^{'}$ and ${{T^3}^{'}}$ are L2 Normalized along the channel, $score \in R^{H_{up} \times W_{up}}$, $H_{up}$ and $W_{up}$ are hyperparameters. 

To provide supervision signals, we construct a binary label $\hat{y} \in \{0,1\}^{H_{up} \times W_{up}}$ by assigning a label of 1 to pixels within the bounding box and 0 to others. The dense matching loss is formulated as a binary cross-entropy (BCE) loss:

\begin{equation}
    \mathcal{L}_{DM}=BCE(Sigmoid(score / \tau), \hat{y}).
\end{equation}
Here $\tau$ represents the temperature coefficient. By leveraging the dense matching loss, SATracker significantly benefits from explicit semantic guidance during training.

\subsection{Prediction Head and Loss}
Following \cite{yan2021learning,yan2021alpha,yang2021siamcorners}, we adopt a fully convolutional approach for localizing target objects using corner estimation. The plain corner head incorporates multiple Conv-BN-ReLU layers that predict the positions of the top-left and bottom-right corners. By calculating the expectation over the distribution of corner probabilities, we can obtain an accurate estimation of the bounding box.

Consequently, our SATracker is trained in an end-to-end manner, supervised with the $L_1$ Loss, the Generalized IoU loss \cite{li2021generalized}, and the Dense Matching loss. The overall loss function is formulated as follows:

\begin{equation}
    \mathcal{L}_{total}=\lambda_{giou}\mathcal{L}_{GIoU}+\lambda_{L_1} \mathcal{L}_1+\lambda_{dm}\mathcal{L}_{DM},
\end{equation}
where $\lambda_{giou},\lambda_{L_1}$ and $\lambda_{dm}$ are hyperparameters.

\section{Experiments}

\subsection{Implementation Details}
Our tracker is built using Python 3.8 and PyTorch 2.0, running on two NVIDIA 3090 GPUs. We first train our model following the procedure of STARK \cite{yan2021learning}. Then we fine-tune it with training splits of TNL2K \cite{wang2021towards}, OTB99 \cite{li2017tracking}, and LaSOT \cite{fan2019lasot}. For optimization, we use ADAM \cite{kingma2014adam} with weight decay of $1 \times 10^{-4}$. The learning rate for backbone is $1 \times 10^{-5}$, and for prediction head, it is $1 \times 10^{-4}$. The fine-tuning process consists of 200 epochs, with learning rates decreasing to one-tenth after 160-th epoch. For hyperparameters, the batch size is 16. The $\lambda_{giou},\lambda_{L_1}$ and $\lambda_{dm}$ are set to 2,5,1 and the upsample size of the dense matching score is $40 \times 40$. We set the temperature parameter $\tau = 0.07$.

\noindent
\textbf{Architectures. }
To better exploit vision cues like classical trackers, we adopt the \textbf{3-stage CvT-21 (Base model)} \cite{wu2021cvt} as the foundational architecture for constructing our hierarchical Vision-Language (VL) tracker. We provide an overview of our backbone architecture in Tab. \ref{t1}, whose input consists of the template ($128 \times 128 \times 3$), the search region ($320 \times 320 \times 3$) and the natural language description ($1 \times 30$). $S$ and $T$ represent for the search region and template, $NL$ denotes the sentence. $H_i$ and $D_i$ is the head number and embedding feature dimension in $i$-th stage. $R_i$ is the feature dimension expansion ratio in the MLP layer. The backbone is loaded with weights of CvT-21 pre-trained on ImageNet. For textual encoder, we utilize the base version of BERT \cite{devlin2018bert} and split it into 3 stages (i.e., [layer0],[layer1:layer4],[layer5:layer11]). The maximum length of the text is 30, including [CLS] and [SEP].
\begin{table}[]\scriptsize
\centering
\begin{tabular}{ccc|c}
\hline
\multicolumn{1}{c|}{\multirow{2}{*}{}} &
  \multicolumn{1}{c|}{\multirow{2}{*}{Layer Name}} &
  \multirow{2}{*}{Output Size} &
  \multirow{2}{*}{SATracker} \\
\multicolumn{1}{c|}{} &
  \multicolumn{1}{c|}{} &
   &
   \\ \hline
\multicolumn{1}{c|}{\multirow{4}{*}{Stage1}} &
  \multicolumn{1}{c|}{CTE} &
  \begin{tabular}[c]{@{}c@{}}$S:   80 \times 80$,\\      $T : 32 \times 32$\end{tabular} &
  \begin{tabular}[c]{@{}c@{}}$7   \times 7$, 64, \\      stride 4\end{tabular} \\ \cline{2-4} 
\multicolumn{1}{c|}{} &
  \multicolumn{1}{c|}{TEM} &
  \begin{tabular}[c]{@{}c@{}}$S:   80 \times 80$, \\      $T : 32 \times 32$\end{tabular} &
  $\left[\begin{array}{c}H_1=1   \\ D_1=64 \\ R_1=4\end{array}\right] \times 1$ \\ \cline{2-4} 
\multicolumn{1}{c|}{} &
  \multicolumn{1}{c|}{BERT} &
  $NL : 1 \times 30$ & [layer0]
   \\ \cline{2-4} 
\multicolumn{1}{c|}{} &
  \multicolumn{1}{c|}{SAM} &
  \begin{tabular}[c]{@{}c@{}}$S:   80 \times 80$,\\      $NL : 1 \times 30$\end{tabular} & VL Fusion $\times 1$
   \\ \hline
\multicolumn{1}{c|}{\multirow{4}{*}{Stage2}} &
  \multicolumn{1}{c|}{CTE} &
  \begin{tabular}[c]{@{}c@{}}$S:   40 \times 40$,\\      $T : 16 \times 16$\end{tabular} &
  \begin{tabular}[c]{@{}c@{}}$3   \times 3$, 192, \\      stride 2\end{tabular} \\ \cline{2-4} 
\multicolumn{1}{c|}{} &
  \multicolumn{1}{c|}{TEM} &
  \begin{tabular}[c]{@{}c@{}}$S:   40 \times 40$,\\      $T : 16 \times 16$\end{tabular} &
  $\left[\begin{array}{c}H_2=3   \\ D_2=192 \\ R_2=4\end{array}\right] \times 4$ \\ \cline{2-4} 
\multicolumn{1}{c|}{} &
  \multicolumn{1}{c|}{BERT} &
  $NL : 1 \times 30$ &[layer1:layer4]
   \\ \cline{2-4} 
\multicolumn{1}{c|}{} &
  \multicolumn{1}{c|}{SAM} &
  \begin{tabular}[c]{@{}c@{}}$S:   40 \times 40$,\\      $NL : 1 \times 30$\end{tabular} & VL Fusion $\times 1$
   \\ \hline
\multicolumn{1}{c|}{\multirow{4}{*}{Stage3}} &
  \multicolumn{1}{c|}{CTE} &
  \begin{tabular}[c]{@{}c@{}}$S:   20 \times 20$,\\      $T : 8 \times 8$\end{tabular} &
  \begin{tabular}[c]{@{}c@{}}$3   \times 3$, 384, \\      stride 2\end{tabular} \\ \cline{2-4} 
\multicolumn{1}{c|}{} &
  \multicolumn{1}{c|}{TEM} &
  \begin{tabular}[c]{@{}c@{}}$S:   20 \times 20$,\\      $T : 8 \times 8$\end{tabular} &
  $\left[\begin{array}{c}H_3=6   \\ D_3=384 \\ R_3=4\end{array}\right] \times 16$ \\ \cline{2-4} 
\multicolumn{1}{c|}{} &
  \multicolumn{1}{c|}{BERT} &
  $NL : 1 \times 30$ &[layer5:layer11]
   \\ \cline{2-4} 
\multicolumn{1}{c|}{} &
  \multicolumn{1}{c|}{SAM} &
  \begin{tabular}[c]{@{}c@{}}$S:   20 \times 20$,\\      $NL : 1 \times 30$\end{tabular} & VL Fusion $\times 1$
   \\ \hline

\end{tabular}
\caption{The backbone architecture of our SATracker.}
\label{t1}
\end{table}

\begin{table*}[]
\centering

\begin{threeparttable}

\begin{tabular}{c|c|c|ccc|cc|cc}
\toprule[0.5mm]
\multirow{2}{*}{Algorithms}&\multirow{2}{*}{Published} & \multirow{2}{*}{Initialize} &  \multicolumn{3}{c|}{TNL2K} & \multicolumn{2}{c|}{OTB99} & \multicolumn{2}{c}{LaSOT}  \\
                            &   &               & SUC   & Norm.PRE  & PRE            & SUC          & PRE         & SUC          & PRE          \\
                            
\midrule[0.3mm]

AutoMatch \cite{zhang2021learn}    &ICCV21                       & BBox     & 47.2   & \multicolumn{1}{c}{-}        & 43.5                  & \multicolumn{2}{c|}{—}    & \multicolumn{1}{c}{58.3}   & 59.9      \\

TrDiMP \cite{wang2021transformer}     &CVPR21                      & BBox    & \textbf{52.3}   & \multicolumn{1}{c}{-}        & \textbf{52.8}                  & \multicolumn{2}{c|}{—}    & \multicolumn{1}{c}{63.9}   & 66.3       \\

TransT \cite{chen2021transformer}     &CVPR21                      & BBox      & 50.7   & \multicolumn{1}{c}{57.1}        & 51.7               & \multicolumn{2}{c|}{—}    & \multicolumn{1}{c}{\underline{64.9}}   & \underline{69.0}       \\

TransInMo \cite{guolearning}    &IJCAI22                       & BBox   & \underline{52.0}   & \underline{58.5}       & \underline{52.7}                   & \multicolumn{2}{c|}{—}    & \multicolumn{1}{c}{\textbf{65.7}}   & \textbf{70.7}       \\

\midrule[0.3mm]

TNLS-III \cite{li2017tracking}   &CVPR17                        & NL+BBox     & \multicolumn{3}{c|}{——}                 & 55.0 & 72.0    & \multicolumn{2}{c}{—}     \\

RTTNLD \cite{feng2020real}    &WACV20                       & NL+BBox    & \multicolumn{3}{c|}{——}              & 61.0 & 79.0    & 35.0 & 35.0     \\

GTI \cite{yang2020grounding}      &TCSVT20                    & NL+BBox   & \multicolumn{3}{c|}{——}                  & 67.2 & 86.3    & 63.1 & 66.5     \\

CapsuleTNL \cite{ma2021capsule}   &MM21                       & NL+BBox     & \multicolumn{3}{c|}{——}                  & 71.1 & 92.4    & 61.5 & 63.3   \\

TNL2K-2 \cite{wang2021towards}     &CVPR21                     & NL+BBox      & 42.0 & 50.0 & 42.0                & 68.0 & 88.0    & 51.0 & 55.0    \\

SNLT \cite{feng2021siamese}       &CVPR21                   & NL+BBox       & 27.6 & - & 41.9                & 66.6 & 80.4    & 54.0 & 57.6   \\

CTRNLT \cite{li2022cross}     &CVPRW22                     & NL+BBox      & 44.0 & 52.0 & 45.0                & 53.0 & 72.0    & 52.0 & 51.0    \\

VLTTT \cite{guodivert}       &NeurIPS22                   & NL+BBox     & 53.1 & 59.3 & 53.3                 & \textbf{76.4} & 93.1    & 67.3 & 71.5    \\

MMTrack\cite{zheng2023towards} & TCSVT23
& NL+BBox & \underline{58.6} &\underline{75.2} &\underline{59.4} &70.5 &91.8 &70.0 &75.7 \\

All-in-One\cite{zhang2023all} &MM23 & NL+BBox &55.3 &- &57.2 &71.0 &93.0 &\underline{71.7} &\underline{78.5}\\


DecoupleTNL\cite{ma2023tracking} &ICCV23 & NL+BBox &56.7 &- &56.0 &73.8 &\textbf{94.8} &71.2 &75.3\\

JointNLT \cite{zhou2023joint}    & CVPR23                      & NL+BBox     & 56.9 & 73.6 & 58.1                 & 65.3 & 85.6    & 60.4 & 63.6    \\

\textbf{SATracker}      & \textbf{Ours}                     & NL+BBox       & \textbf{61.6} & \textbf{78.4} & \textbf{64.4}           & \underline{74.2} & \underline{94.3}    & \textbf{72.4} & \textbf{79.6}   \\

\bottomrule[0.5mm]
\end{tabular}

\caption{Success (SUC), Precision (PRE), and Normalized Precision (Norm.PRE) of different trackers on the TNL2K, OTB99, and LaSOT. The best and second-best results are marked in \textbf{bold} and \underline{underline} accordingly. BBox and NL represent the Bounding Box and Natural Language, respectively.(\%)}
\label{t2}

\end{threeparttable}
\end{table*}

\noindent
\textbf{Datasets and metrics. }
We evaluate the effectiveness of our approach on popular benchmarks that are specifically designed for VL tracking, i.e., TNL2K and OTB99. Additionally, we evaluate our method on LaSOT, a long-term visual tracking benchmark that provides natural language descriptions. All these datasets adopt success, precision and normalized precision to measure the performance, which are commonly adopted in the field of VL tracking.

\subsection{Comparison with the State-of-the-art Trackers}

\begin{table}[]\footnotesize
\centering
\begin{tabular}{ccccc}
\toprule[0.5mm]
\multirow{2}{*}{} & \multirow{2}{*}{FLOPs (G)} & \multirow{2}{*}{Params (M)} & \multicolumn{2}{c}{SUC (\%)} \\
                 &      &       & TNL2K & LaSOT \\
\midrule[0.3mm]
Ours & 28.8 & 131.3 & \textbf{61.6}  & \textbf{72.4}  \\
MMTrack          & - & 176.9 & \underline{58.6}  & \underline{70.0}  \\
JointNLT         & 42   & 153   & 56.9  & 60.4  \\
VLTTT            & 14.7 & 100.9 & 53.1  & 67.3  \\
SNLT             & 66.5 & 149.5 & 27.6  & 54.0  \\
\bottomrule[0.5mm]
\end{tabular}
\caption{For fair comparisons, we provide efficiency metrics such as FLOPs and Params, eliminating influences of hardware devices and optimization techniques.}
\label{t3}
\end{table}

We present a comprehensive comparison of SATracker with SOTA trackers in Tab. \ref{t2}, where our method exceeds others on VL datasets (i.e., TNL2K and OTB99). Moreover, it achieves excellent results on long-term tracking dataset LaSOT, without any design for online updating.

\noindent
\textbf{TNL2K. }
TNL2K is a newly introduced benchmark designed for VL tracking that offers a large-scale dataset containing 2,000 video sequences. It stands out due to several desirable features: high quality, adversarial samples and significant appearance variation. As shown in Tab. \ref{t2}, our method surpasses them by a large margin: we outperform MMTrack by 3.0\%, 3.2\%, and 5.0\% in SUC, Norm.PRE, and PRE, respectively. Moreover, our efficient tracker achieves these exceptional results while utilizing fewer FLOPs and Params compared to publicly available models in Tab. \ref{t3}, except for VLTTT that utilizes the Neural Architecture Search \cite{elsken2019neural} technique to significantly reduces number of parameters.

\noindent
\textbf{OTB99. }
The OTB99 dataset stands as the pioneering VL tracking dataset, which is relatively small-scale with only 51 videos for training and 48 videos for testing. While SATracker falls behind VLTTT, our method still ranks as the second among all VL trackers in Tab. \ref{t2}. The reason is that VLTTT is specifically trained and evaluated on each datasets, with the best results reported. In contrast, our model is trained and evaluated across all datasets \textbf{with same model weights and hyper-parameters}, following the standard protocol of tracking. Consequently, our approach exhibits better generalization ability.

\noindent
\textbf{LaSOT. } 
LaSOT, a comprehensive benchmark consists of long sequences for visual tracking, whose primary challenge lies in the robustness of long-term tracking. It comprises 280 test sequences, each averaging 2500 frames. Among all tracker, our method stands out as the top-performing method, surpassing All-in-one by 0.7\% and 1.1\% in terms of the SUC and PRE, respectively.

\subsection{Ablation Study and Analysis}
\begin{table*}
\centering
\begin{tabular}{l|ccc|ccc|cc}
\toprule[0.5mm]
\multirow{2}{*}{Experiment Setup} &
  \multirow{2}{*}{TEM} &
  \multirow{2}{*}{SAM} &
  \multirow{2}{*}{DM Loss} &
  \multicolumn{3}{c|}{TNL2K} &
  \multicolumn{2}{c}{OTB99} \\
 &         &         &         & SUC  &Norm.PRE & PRE & SUC  & PRE  \\
\midrule[0.3mm]
\ding{172} (Baseline) &  -       &  -       &    -     &45.2  &  58.7     &42.9          & 42.7    & 52.1         \\
\ding{173} (w/. TEM) & $\surd$ &  -       &   -      & 59.8 & 76.7        &  61.1                &   66.4        &85.2      \\
\ding{174} (w/o. DM) &  $\surd$ & $\surd$ &   -              & 61.0 & 77.8  & 63.5         & 70.1 & 90.3       \\
\ding{175} (\textbf{SATracker}) & $\surd$ & $\surd$ & $\surd$ &  \textbf{61.6} & \textbf{78.4} & \textbf{64.4}        &\textbf{74.2}          &  \textbf{94.3} \\
\bottomrule[0.5mm]
\end{tabular}
\caption{Ablation of main components of SATracker.(\%).}
\label{t4}
\end{table*}

We first perform ablation studies to thoroughly analyze the impact of main components within our model. Then, explorations of key designs of the synchronous learning backbone and each module within it are provided. To ensure a proper evaluation against challenges of this task, all experiments are conducted on two benchmark datasets specifically designed for VL tracking: TNL2K and OTB99.

\noindent
\textbf{Ablation of Main Components. }
To verify the contribution of each component, we conduct ablation studies on four variants of our model. The experimental results of these variants on all datasets are presented in Tab. \ref{t4}.

\begin{itemize}
    \item \textbf{Setting \ding{172}} (Baseline) adopts the 3-stage CvT as the backbone, which only utilize layers of self attention to extract visual features. In this variant, we remove the SAM while retain the corner head. To further improve its performance, we leverage the ECA and FCA of the powerful tracker TransT \cite{chen2021transformer} as its neck module. 
    \item \textbf{Setting \ding{173}} (w/. TEM) introduces asymmetrical cross attention upon the \textbf{Baseline}. We remove the neck module since it already has the cross-correlation design. This demonstrates the effectiveness of direct interactions between the template and search branches during feature extraction, leading to improved performance.
    \item \textbf{Setting \ding{174}} (w/o. DM) incorporates the textual modality to enhance pure visual tracker (\textbf{Setting \ding{173}}). Via SAM, we inject high-level semantics into the search feature to better perceive the target from both modalities.
    \item \textbf{Setting \ding{175}} (SATracker) further utilizes the DM Loss on the base of \textbf{Setting \ding{174}}, unleashing the power of VL representation learning for VL tracking.
\end{itemize}

\begin{figure}
\centering 
\includegraphics[width = 8cm]{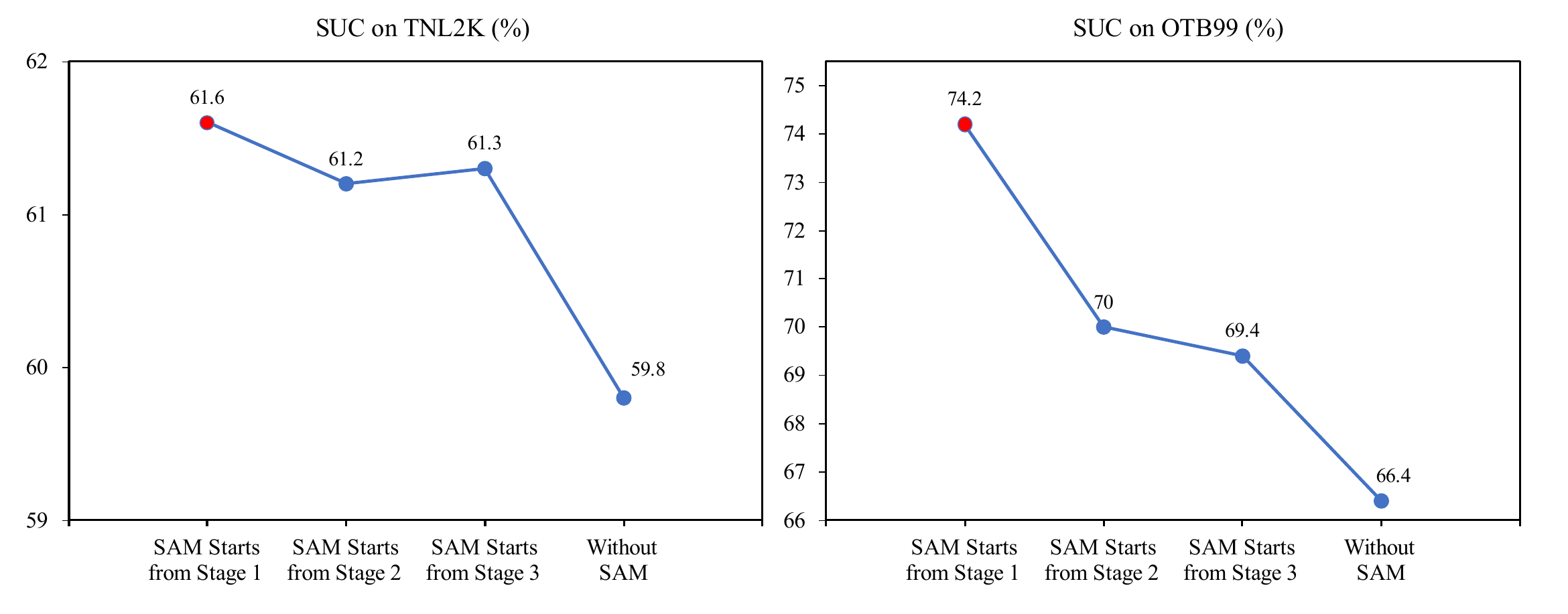}
\caption{The impact of the starting stage of SAM.}
\label{f5}
\end{figure}

\noindent
\textbf{Impact of Synchronous Learning Backbone. }
We also conduct experiments to verify our motivations for constructing a Synchronous Learning Backbone (SLB), as detailed in Tab. \ref{t5}. In the `asynchronous' variant, we utilize the textual features pre-extracted by BERT to initiate interaction with visual features. However, the direct use of well-learned high-level textual features hinders the synchronous learning with visual features, which naturally progress from low-level to high-level. In the `even-split' variant, 12-layers BERT are evenly split into 3 stages. Additionally, we investigate the starting stage of SAM to validate the necessity of VL fusion at each stage, which are depicted in Fig. \ref{f5}. The place order of TEM and SAM is also explored in the variant `reversed', where we place SAM before TEM in each stage. Above-mentioned results clearly demonstrate that enabling synchronous learning and interactions between vision and language from the beginning is crucial for effectively exploring target-relevant semantics.

\begin{table}[] \small
\centering
\begin{tabular}{c|c|cc|cc}
\toprule[0.5mm]
\multirow{2}{*}{Setting} & \multirow{2}{*}{Module} & \multicolumn{2}{c|}{TNL2K} & \multicolumn{2}{c}{OTB99} \\

               &                      & SUC & PRE & SUC & PRE \\
\midrule[0.3mm]
asynchronous  & \multirow{3}{*}{SLB} & 60.5    & 62.3  &   68.6    &87.3     \\
reversed    &                      &61.4     &  63.7     &70.3   &89.9     \\
even-split    &                      &61.4     &  64.1     &70.1   &89.7     \\
               \midrule[0.3mm]
self attention & \multirow{2}{*}{TEM} & 52.9    & 53.0  &   52.2    &65.5     \\
symmetrical    &                      &60.9     &  63.3     &69.1   &89.1     \\
\midrule[0.3mm]
w/o. SAM       & \multirow{3}{*}{SAM} &59.8     &61.1    & 66.4     & 85.2    \\
ModaMixer      &                      & 61.2    & 63.6   & 69.1     & 88.3    \\
w/o. update    &                      & 61.4    &  63.7  & 69.7     &89.6     \\
\midrule[0.3mm]
\textbf{ours}           &  \textbf{SATracker}                    & \textbf{61.6}    & \textbf{64.4}   & \textbf{74.2}     &\textbf{94.3}    \\
\bottomrule[0.5mm]
\end{tabular}
\caption{The influence of some key designs.(\%)}
\label{t5}
\end{table}

\noindent
\textbf{Impact of Asymmetrical Cross Attention. }
To further investigate the design of the Target Enhance Modules (TEM), we explore two variants: `self attention' and `symmetrical'. These variants, along with SATracker, differ in the design of the feature extraction within TEM: pure self-attention, symmetrical cross-attention, and asymmetrical cross-attention. As illustrated in Tab. \ref{t5}, the performance gap between `self attention' and `symmetrical' demonstrates the effectiveness of cross-correlation operation between the template and search during feature extraction. When comparing the `symmetrical' variant with ours, we observe that alleviating negative influences of noise from the search to the template yields improved performance.

\noindent
\textbf{Impact of VL Fusion Design. }
The integration of natural language description has significantly enhanced the performance of classical visual trackers.  The key aspect lies in effectively leveraging both visual and textual modalities. To verify the effectiveness of our VL fusion design, we introduce two variants: `w/o. SAM' and `ModaMixer'. The `w/o. SAM' variant represents a pure visual tracker that solely incorporates TEM, while `ModaMixer' replaces our fusion design with the fusion design utilized in VLTTT \cite{guodivert}. The performance comparison in Tab. \ref{t5} between `w/o. SAM' and `ours' clearly demonstrates the necessity of incorporating textual modality. Our VL fusion design outperforms the `ModaMixer', which claims to be a universal and effective fusion design.

\noindent
\textbf{Impact of Updating Textual Features. }
To tackle challenges posed by variations in target appearance, we improve the adaptability of our model by updating textual features. The benefits of this enhancement are evident in Tab. \ref{t5}, where we observe a noticeable performance gap between the `w/o. update' and `ours'. The update of textual features plays a crucial role in enhancing the robustness of our model, enabling it to handle significant appearance changes.

\subsection{Visualization Results}
\begin{figure}
\centering 
\includegraphics[width = 8cm]{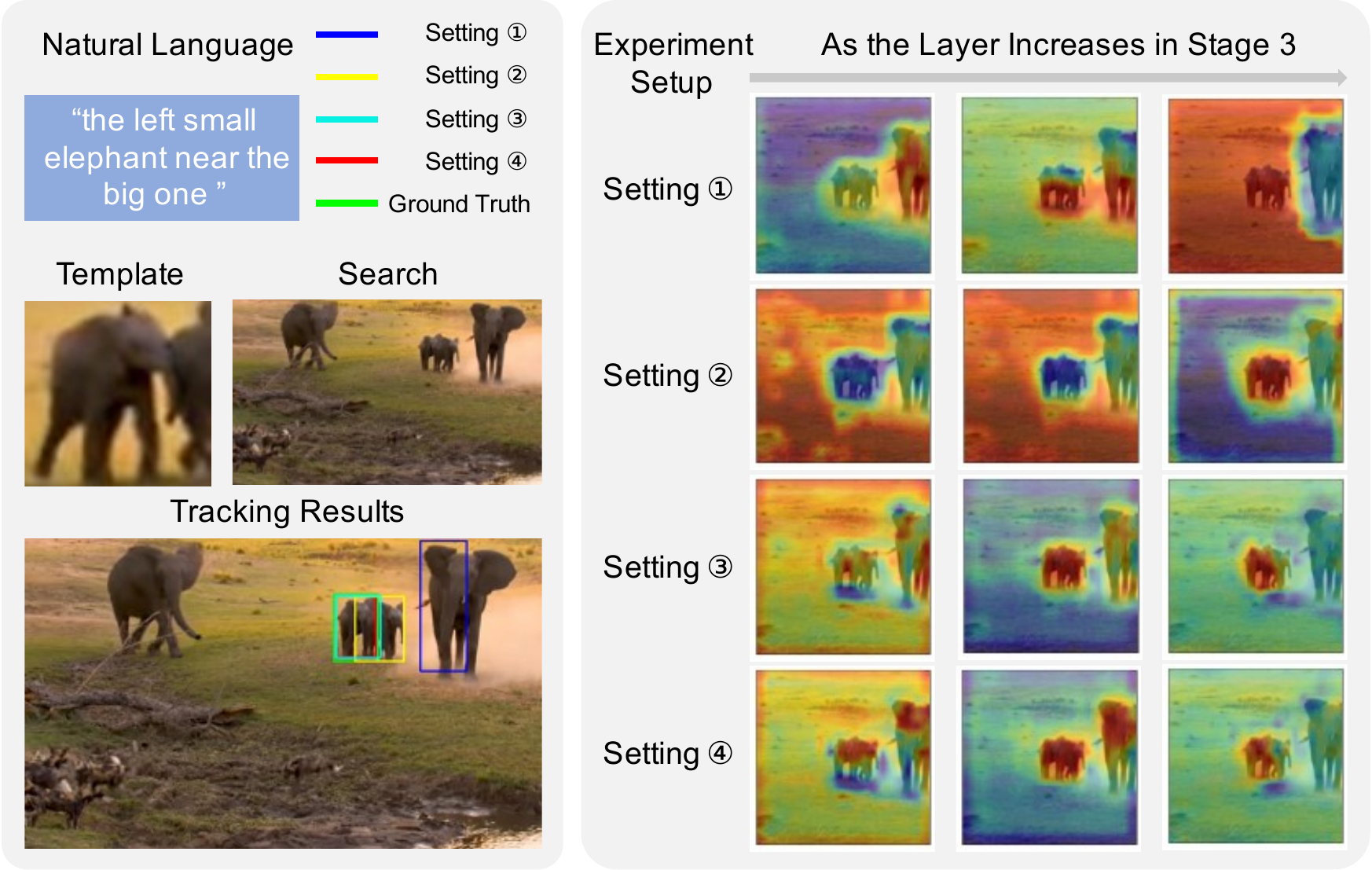}
\caption{Qualitative ablations of main components.}
\label{f6}
\end{figure}
As depicted in Fig. \ref{f6}, we present the visualized tracking results of four variants from the \textbf{Ablation of Main Components} part, and corresponding response maps from Stage 3 on a challenging video from TNL2K dataset. This video involves difficulties such as \textit{semantic understanding and distractions from similar objects}. Notably, as the layer increases, all variants, except for setting \ding{172}, are capable of locating the target to some extent. However, only setting \ding{174} and setting \ding{175} achieve precise and accurate target localization. This observation validates that the Target Enhance Module (TEM) allows us to perceive the target effectively, while the Semantic Aware Module (SAM) takes it a step further by enabling a deeper understanding of semantics, thereby mitigating ambiguity and distractions. Furthermore, Dense Matching (DM) loss accelerates the process of focusing on the target and ensures superior performance. 

Moreover, we conduct \textbf{more qualitative comparisons of SATracker with other SOTA VL trackers on challenging sequences} in TNL2K and OTB99. Our method is also discussed for its \textbf{limitations}, which potentially leads to slightly inferior performance on LaSOT. For further visualizations and analyses, please refer to our supplementary material.

\section{Conclusion}
Our SATracker introduces the first Synchronous Learning Backbone for VL tracking. It mainly consists of Target Enhance Modules (TEM) and Semantic Aware Modules (SAM), progressively capturing relations between visual and textual modalities. Furthermore, we propose a dense matching loss to optimize VL learning. Extensive experiments validate the effectiveness and generalization of our approach.

{
    \small
    \bibliographystyle{ieeenat_fullname}
    \bibliography{main}
}


\end{document}